\documentclass[twoside,11pt]{article}

\usepackage{jmlr2e}

\usepackage[final]{microtype}
\usepackage{booktabs}
\usepackage{multirow}
\usepackage{array}
\usepackage{dblfloatfix}
\usepackage{siunitx}
\usepackage{color,soul}
\definecolor{mypink1}{rgb}{1.00,0.97,0.93}
\sethlcolor{mypink1}
\let\oldcite\cite
\renewcommand{\cite}[1]{\mbox{\oldcite{#1}}}
\usepackage{url}

\AtBeginDocument{%
  \setlength\abovedisplayskip{1.35ex}
  \setlength\belowdisplayskip{1.35ex}}

\usepackage{subfigure}
\usepackage{caption}

\firstpageno{1}
\newcommand\Mark[1]{\textsuperscript#1}

\begin{document}

\title{Domain Adaptation for Rare Classes Augmented with Synthetic Samples}

\begingroup
\centering
{\LARGE\bfseries Domain Adaptation for Rare Classes Augmented with Synthetic Samples \\[1.4em]
\large\mdseries Tuhin Das\Mark{1}, Robert-Jan Bruintjes\Mark{1}, Attila Lengyel\Mark{1}, Jan van Gemert\Mark{1} \\and Sara Beery\Mark{2}}\\[1.3em]
\begin{tabular}{*{2}{>{\centering}p{.40\textwidth}}}
\Mark{1}Computer Vision Lab & \Mark{2} Computational Vision Lab \tabularnewline
Delft University of Technology & California Institute of Technology \tabularnewline
% \url{{asdasldkasmd}} & \url{email2} 
\end{tabular}\par
\endgroup

\vspace{2.5em}

\begin{abstract}%
To alleviate lower classification performance on rare classes in imbalanced datasets, a possible solution is to augment the underrepresented classes with synthetic samples.
Domain adaptation can be incorporated in a classifier to decrease the domain discrepancy between real and synthetic samples.
While domain adaptation is generally applied on completely synthetic source domains and real target domains, we explore how domain adaptation can be applied when only a single rare class is augmented with simulated samples.
As a testbed, we use a camera trap animal dataset with a rare \textit{deer} class, which is augmented with synthetic deer samples.
We adapt existing domain adaptation methods to two new methods for the single rare class setting:
\textit{DeerDANN}, based on the Domain-Adversarial Neural Network (DANN), and \textit{DeerCORAL}, based on deep correlation alignment (Deep CORAL) architectures.
Experiments show that DeerDANN has the highest improvement in deer classification accuracy of 24.0\% versus 22.4\% improvement of DeerCORAL when compared to the baseline.
Further, both methods require fewer than 10k synthetic samples, as used by the baseline, to achieve these higher accuracies.
DeerCORAL requires the least number of synthetic samples (2k deer), followed by DeerDANN (8k deer).
\end{abstract}

\section{Introduction}

Computer vision models generally perform well on datasets where each class is well-represented~\cite{deng2009}.
However, real-world datasets often have long-tailed distributions~\cite{horn2017,zhu2014,salakhutdinov2011}, which can lead to class imbalances.
As a result, the performance of computer vision algorithms on underrepresented classes can be inferior to the performance on well-represented classes~\cite{horn2017, beery2018}.
In some domains, recognition of rare classes is especially important, e.g. for the monitoring of rare animals in camera trap datasets, which motivates the investigation of methods for improving rare-class performance.
A possible method to reduce the performance discrepancy between imbalanced classes is to oversample the rare classes with new samples~\cite{buda2018}.
As finding new samples of rare classes can be difficult, synthetic samples can be used instead.
Generating synthetic samples of rare classes is especially attractive as computer vision algorithms have been shown to perform well on real test sets when trained on fully synthetic datasets~\cite{chen2017, gaidon2016, marin2010, varol2017} or on synthetically augmented datasets~\cite{fridadar2018, pishchulin2012, pishchulin2011}.

As a real-world testbed, in this work we focus on camera trap data containing images of animals captured by heat or motion-activated passive monitoring cameras.
These cameras are widely used to monitor biodiversity and animal behavior, and to measure the efficacy of conservation actions.
Camera traps collect enormous datasets, up to millions of images from a single network of cameras in a single season.
It is inhibitively time-consuming for ecologists to manually label the species seen in the data, thus automated classification of animal species is needed to match the speed of data collection.

The distribution of species in camera trap datasets is long-tailed~\cite{beery2018, norouzzadeh2018, beery2019c, beery2019b, beery2020b, beery2021a}, mimicking the distribution of species in the natural world.
In~\cite{beery2020}, the Caltech Camera Traps (CCT) dataset~\cite{beeryCCT, beery2018}, which contains a rare \textit{deer} class, was augmented with synthetic deer images generated with 3D game engines to reduce the class imbalance and increase the pose and location diversity of the rare class samples.
While~\cite{beery2020} found that the synthetically augmented training set led to considerably reduced classification errors for the deer class, the performance was still low - average precision for the rare class maxed out at 66\% using 100k synthetic samples. It also appeared that there was no overlap between the deep network representations of the real deer and the synthetic deer.
As the synthetic deer are not photorealistic and do not have realistic image statistics, there is a domain discrepancy between the real and synthetic deer.
Our objective is to minimize the domain discrepancy between real and synthetic deer to improve recognition of the rare deer class. 

In this work, we propose using domain adaptation methods to minimize the domain discrepancy between real and synthetic deer.
Domain adaptation techniques are used to create a shared feature space between a source and a target domain, such that an algorithm trained on the source domain can be applied directly to the target domain~\cite{pan2010}.
Domain adaptation has been successfully used to reduce domain discrepancy between real and synthetic samples when training on a synthetic source domain and testing on a real target domain~\cite{atapour2018, dundar2018, 2hong2018, sankaranarayanan2018, wang2019}. 
In our case, the source and target domain are not the training and test set, but the sets of synthetic deer and real deer respectively, both of which are contained in the training set.
Additionally, other animal classes are included in the dataset, represented by real samples only, which need to be classified accurately as well.

We explore two methods of applying domain adaptation on synthetic and real deer samples.
The first method, which we refer to as DeerDANN, is based on the Domain-Adversarial Neural Network (DANN)~\mbox{\cite{ganin2016}} and incorporates a domain discriminator in a classifier with the task to guess which domain a sample belongs to.
While DANNs originally apply the discriminator to samples from each class, we modify the network such that the discriminator only receives deer samples and guesses whether a sample is a real deer or synthetic deer.
This modification removes the unnecessary domain confusion loss of non-deer samples.
By maximizing the discriminator loss and minimizing the classification loss, the classifier learns to extract domain-invariant features for the deer class, while the features from all classes retain class discriminability.

The second method, which we refer to as DeerCORAL, incorporates a correlation alignment (CORAL) loss~\mbox{\cite{sun2016}} in a classifier, as in Deep CORAL architectures~\mbox{\cite{sun2016_2}}.
The CORAL loss represents the distance between the second-order statistics of the source and target domain.
By minimizing the CORAL loss and classification loss, the domain discrepancy is reduced while the features retain class discriminability.
As synthetic samples are only available for the rare class, we organise the real and synthetic data as source and target domains in a different manner than for the original DANN and Deep CORAL methods.
Fig. ~\mbox{\ref{fig:architectures}} presents the schematic architectures of our methods as well as the new domain organisation.

The main contribution of this work is that we adapt the DANN and Deep CORAL domain adaptation techniques to our setting, where synthetic target domain data is only available for a single rare class.
As a result, we significantly improve the classification accuracy of the rare deer class in the CCT dataset~\mbox{\cite{beeryCCT, beery2018}} when compared to a baseline model from~\mbox{\cite{beery2020}}, without notably reducing performance on the other classes in the dataset. Additionally, we show that the number of synthetic samples needed can be greatly reduced when applying domain adaptation on the synthetic samples.

\section{Related Work}

\paragraph{Domain Adaptation}

Domain adaptation is used to make models learn transferable representations from a source dataset, to enable making inference on a target dataset directly~\cite{wang2018}.
Some methods align the distributions of the source and target domains by minimizing some distance between the source and target distributions, such as the Maximum Mean Discrepancy (MMD)~\cite{longDeep2017, tzengDeep2014, yanMind2017} or the correlation alignment (CORAL)~\cite{peng2017, sun2016, sun2016_2}.
Domain-Adversarial Neural Networks (DANNs)~\cite{ganin2015, ganin2016} and Adversarial Discriminative Domain Adaptation (ADDA)~\cite{tzeng2017} confuse a discriminator that classifies sample features to the source or target domain to generate domain-invariant features.
Other methods apply Generative Adversarial Networks (GANs) to generate samples that are similar to target samples while retaining annotations from source samples~\cite{bousmalis2017, isolaImage2017, liu2016}.
A final class of methods applies encoder-decoders to learn domain-invariant feature representations and to perform data \mbox{reconstruction}~\cite{ghifary2016, zhuangSupervised2015}.

Generally, domain adaptation methods are applied between a source training set and a target test set, where examples of all classes of interest are present in both sets.
Our work differs in that respect, as we attempt to apply domain adaptation on real and simulated samples within a single class, without affecting performance on the other classes.

\paragraph{Domain Adaptation for Synthetic Data}

Domain adaptation has often been applied for transfer learning between a synthetic source domain and a real target domain.
An autoencoder-based approach for lane detection has been used in~\cite{garnett2020}, where an autoencoder was trained in unsupervised and semi-supervised settings on synthetically generated road images. 
Deep Generative Correlation Alignment Networks (DGCANs)~\cite{peng2017} can generate new synthetic samples by combining a 3D CAD synthetic domain and a real domain for improving object detection.
Pixel-level domain adaptation on simulated objects using a discriminator and a generator has been used to train a robotic arm to grasp real-world objects~\cite{bousmalis2018}.
A GAN-based image-to-image translation method called CycleGAN~\cite{zhu2020} has been used to transform synthetic crowd images into realistic images to improve crowd counting~\cite{wang2019}.
Style transfer and adversarial learning have been combined to learn monocular depth estimation from synthetic images~\cite{atapour2018}.
Finally, the approach of~\cite{2hong2018} improves face recognition with only single samples available per person by generating synthetic poses of existing samples and performing adversarial domain adaptation on the synthetic data. 

Many of these methods~\cite{2hong2018, atapour2018, bousmalis2018, garnett2020} utilize a discriminator, originating from DANNs, to introduce domain confusion between real and synthetic samples.
While DANNs fall into the supervised domain adaptation category, DGCANs~\cite{peng2017} minimize CORAL loss between real and synthetic samples, which is an unsupervised domain adaptation technique.
To test both supervised and unsupervised methods in our setting, we investigate how DANNs and CORAL loss can be applied to synthetic samples of a single rare class.

\section{Dataset}

We use the Caltech Camera Traps (CCT) dataset~\cite{beery2018} as training and test data.
The original CCT dataset contains 243,187 images of 30 different animal classes captured by heat- or motion-triggered cameras across 140 wildlife locations.
We split the dataset following the CCT-20 data split as described in~\cite{beery2018}, containing 57,868 images with accompanying bounding boxes of 15 animal classes across 20 camera locations.

The CCT-20 split was originally proposed to evaluate the generalization of camera trap classification and detection models on test sets with different image statistics.
To achieve different training and test distributions, the CCT-20 data set is split across camera locations using two partitions: 
\textbf{cis} images are from camera locations seen in the training set and \textbf{trans} images are from locations unseen in the training set to test generalization to new locations.

\begin{figure}[!htb]
   \centering
    \vspace{-1.5ex}
   \subfigure[CCT-20 training]{\label{fig:train_frequencies}\includegraphics[height=0.35\textwidth]{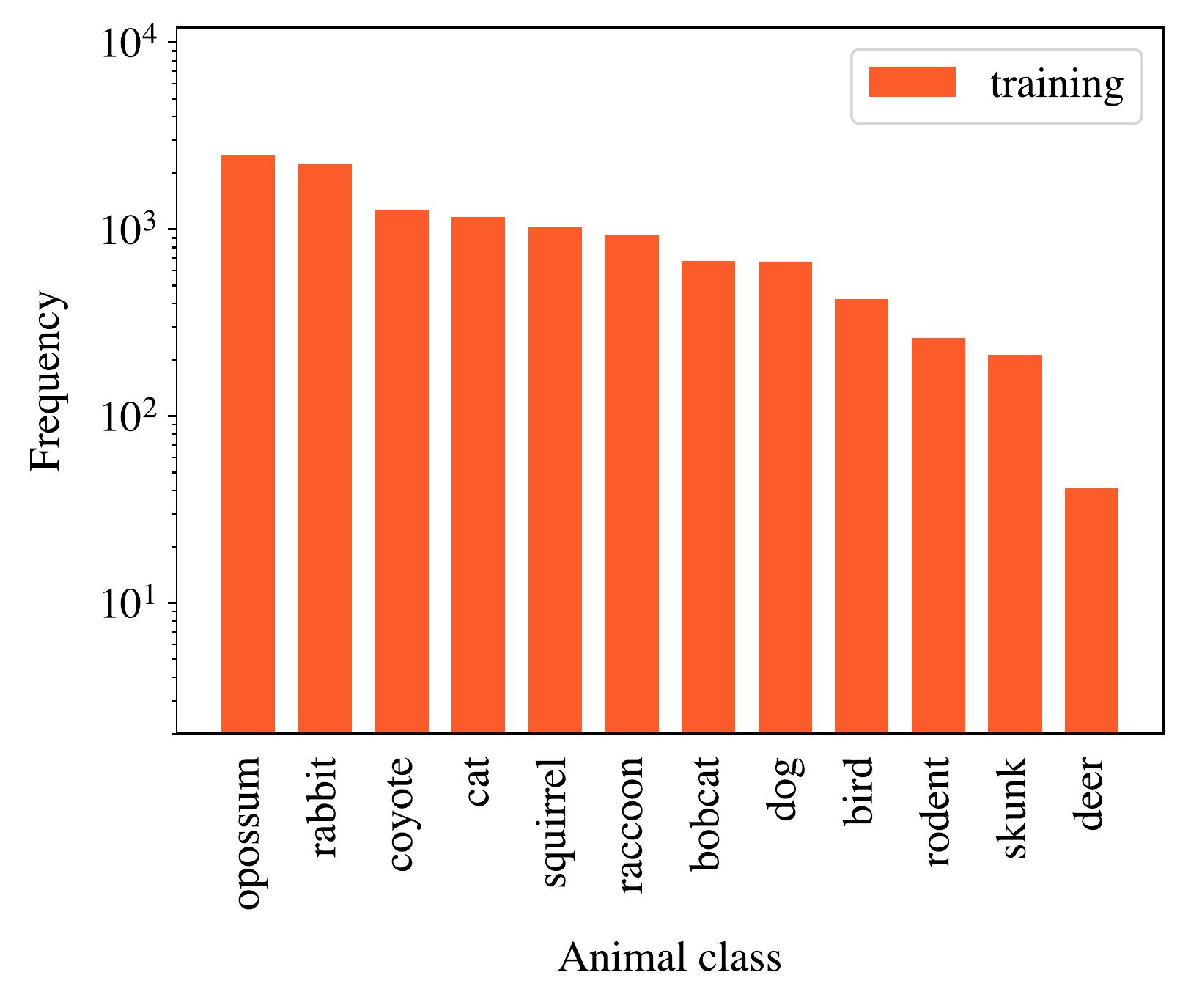}}
   \subfigure[CCT-20 cis and trans+ test]{\label{fig:other_frequencies}\includegraphics[height=0.35\textwidth]{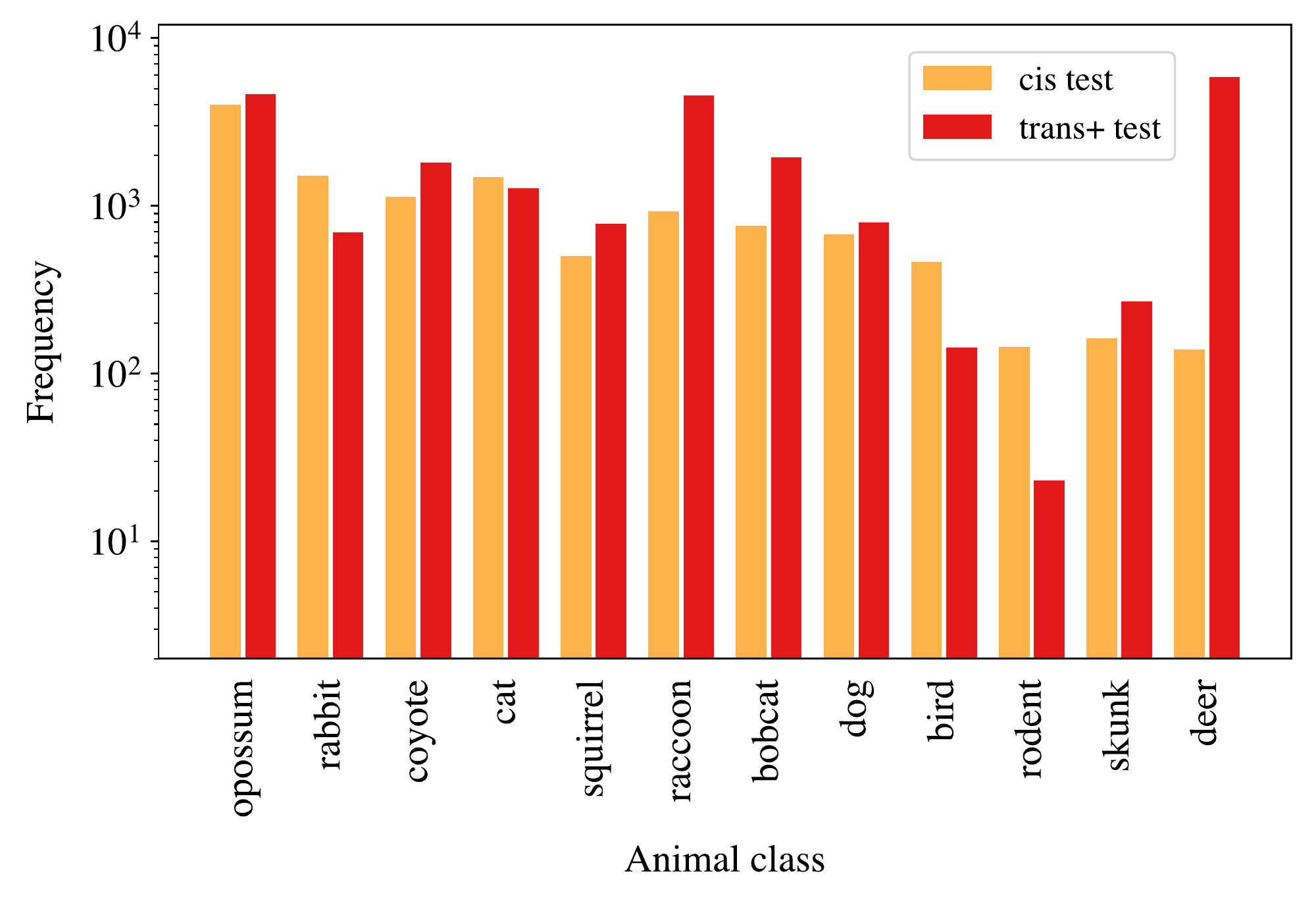}}

   \caption{The animal class frequencies in the CCT-20 training set (a) and test sets (b).
   The training set shows the long-tailed nature of the animal class distribution, in which the deer class is the rarest class with only 41 samples.}\label{fig:frequencies}
\end{figure}

The CCT-20 data set is split into five subsets: training, cis validation, cis test, trans validation, and trans test.
The CCT-20 training set clearly shows the long-tailed distribution of animal classes, with the deer class as the rarest class in the training data with only 41 samples (see Fig.~\ref{fig:train_frequencies}).

As the trans sets do not contain any deer samples in the CCT-20 split, we augment the trans sets with deer samples from other CCT locations to become \textbf{trans+} sets (see Fig.~\ref{fig:other_frequencies}), as in~\cite{beery2020}.
Following~\cite{beery2020}, we also remove the \textit{badger}, \textit{fox}, \textit{empty} and \textit{car} classes from the data sets to retain only images containing animals (focusing on classification as opposed to detection), and to isolate the deer class as the single rare class to prototype our approach.

\begin{figure}[!htb]
   \centering

   \subfigure[Real day]{\includegraphics[height=0.20\textwidth]{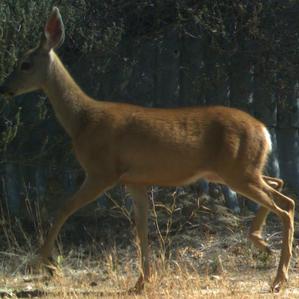}}
   \subfigure[Real night]{\includegraphics[height=0.20\textwidth]{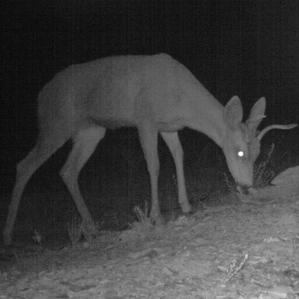}}
   \subfigure[Simulated day]{\includegraphics[height=0.20\textwidth]{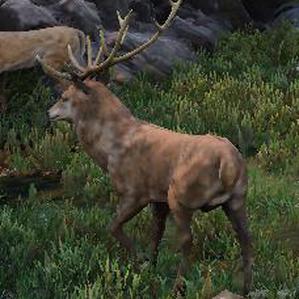}}
   \subfigure[Simulated night]{\includegraphics[height=0.20\textwidth]{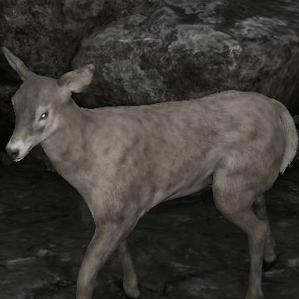}}

   \caption{Real deer images from the CCT-20 training set (a, b) and synthetic deer images generated with the Unity 3D game engine (c, d) used to augment the rare deer class.}\label{fig:deer_matrix}
\end{figure}

As synthetic data we use the simulated deer images that were generated for the CCT-20 dataset in~\cite{beery2020}.
The synthetic deer are created using the Unity 3D game development engine and have accompanying bounding boxes.
To improve the classification performance of the deer class, the simulations are varied in models, pose, lighting, and day or night-time rendering.
Fig.~\ref{fig:deer_matrix} shows examples of real and synthetic deer images.

\section{Methods}

We adapt two domain adaptation techniques, DANN~\cite{ganin2016} and Deep CORAL~\cite{sun2016_2}, to our setting where synthetic data is available for only a single rare class.
As our source and target domains are subsets of the training set, with no target data in the test set, we set up the domains differently than with the standard DANN and Deep CORAL methods.
Here we describe our proposed adaptations and how we organise the data as source and target domains.

\subsection{DeerDANN}

The first method we consider uses a deep network with a feature extractor $F$, domain discriminator $D$, and label classifier $C$, similar to the DANN architecture~\cite{ganin2016}.
The discriminator is given the task to classify all deer features from $F$ as a source (synthetic) or target (real) deer sample.
The source domain $S$ contains the full CCT-20 training set and is extended with synthetic deer.
The target domain $T$ contains the 41 real deer oversampled 50 times to end up with 2050 real deer.
% $T$ is constructed in this manner to resemble $S$ in class priors, but with the features of real deer instead of synthetic deer.

During training, samples from both domains go through $F$ to generate a feature vector $f$.
All deer features $f$ are sent to $D$ and $D$ guesses which domain each deer $f$ belongs to, leading to a domain confusion loss $\mathcal{L}_D$ for source and target deer samples.
$C$ classifies all $f$ coming from $S$, leading to a classification loss $\mathcal{L}_C$.
The network is trained end-to-end using the composite loss
\begin{equation}
   \mathcal{L}_{\text{sum}} = \sum_{i \in S} \mathcal{L}_C^i + \sum_{i \in \delta (S \cup T)} \mathcal{L}_D^i,
\end{equation}

\begin{figure}[!t]
   \centering
   \includegraphics[width=\textwidth]{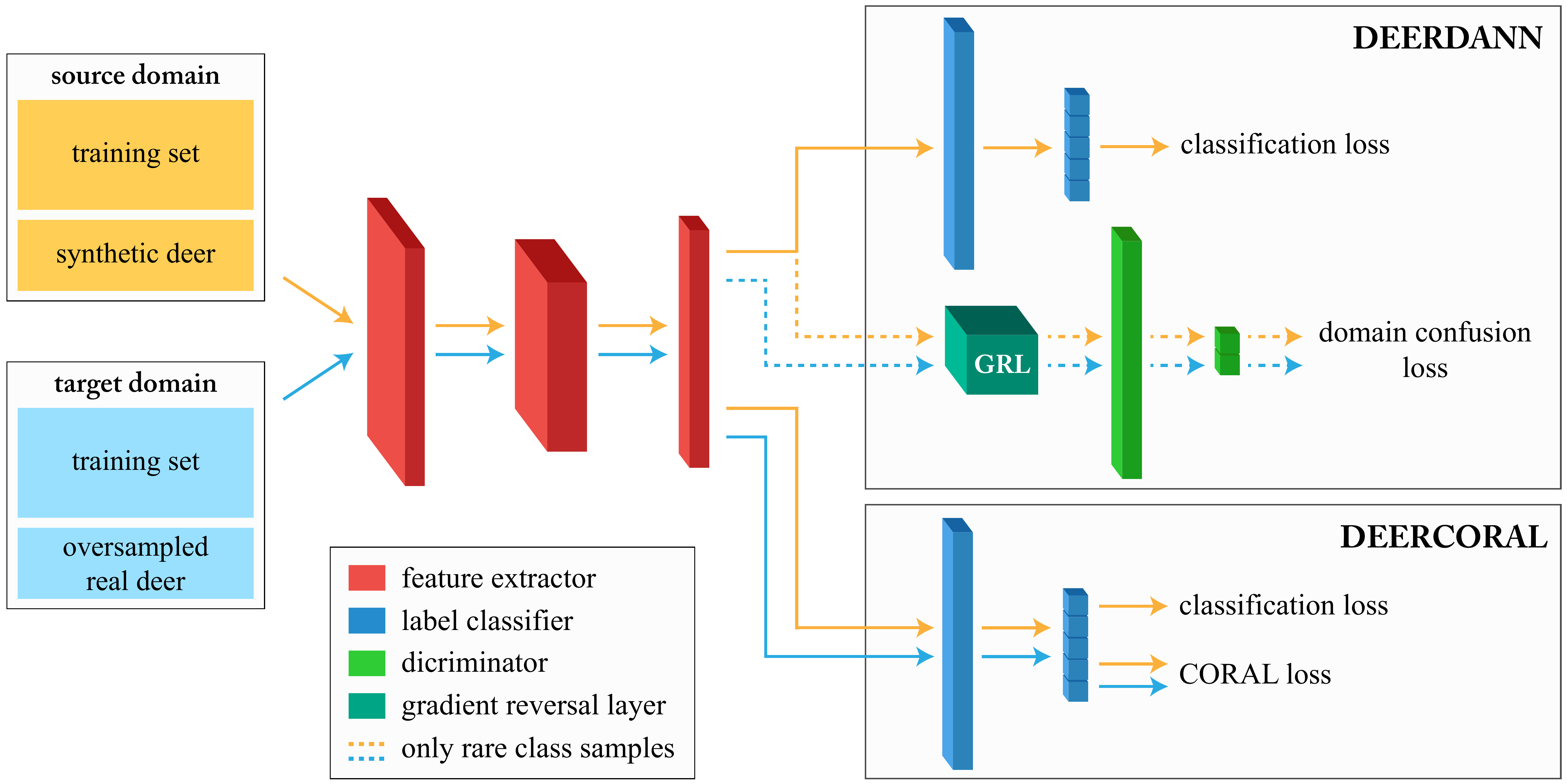}
   \caption{Proposed architectures for adapting either DANN~\cite{ganin2016} or Deep CORAL~\cite{sun2016_2} to synthetic samples of a single rare class. \textbf{(a) DeerDANN}. A label classifier is used to predict the category of source examples only, while the discriminator is used to predict the domain of deer samples. Classification loss is minimized, while domain confusion loss is maximized by the Gradient Reversal Layer (GRL) to generate domain-invariant deer features. Note that the target domain does not need to contain the training set for DeerDANN. \textbf{(b) DeerCORAL}. A feature extractor and label classifier are enhanced with a CORAL loss term, computed with second-order statistics of the source and target domain. CORAL loss and classification loss of source features are both minimized.}
   \label{fig:architectures}
\end{figure}

\noindent where $\delta(\cdot)$ represents all deer features from a domain.
The parameters of $F$ and $C$ are optimized to minimize $\mathcal{L}_C$.
By inserting a Gradient Reversal Layer (GRL)~\cite{ganin2015} between $F$ and $D$, $F$ is updated to maximize $\mathcal{L}_D$ while $D$ is updated to minimize $\mathcal{L}_D$.
As a result the network adversarially learns to generate deer features that are domain-invariant and are discriminative from other classes.
Fig.~\ref{fig:architectures} shows the proposed architecture of DeerDANN.
Note that Fig.~\ref{fig:architectures} includes the training set in the target domain for simplicity, which does not make a difference for training DeerDANN.

\subsection{DeerCORAL}

The second method we consider, DeerCORAL, utilizes correlation alignment like Deep CORAL architectures \cite{sun2016_2} by incorporating a CORAL loss that represents the distance between the second-order statistics of the source and target domain~\cite{sun2016}.
The CORAL loss is computed over the classification logits of a deep network with
\begin{equation}
   \label{CORAL_loss}
   \mathcal{L}_{\text{CORAL}} = \frac{1}{4d^2} \| C_S - C_T\| ^2 _F,
\end{equation}

\noindent where $\|\cdot \|^2_F$ denotes the squared Frobenius norm, $C_S$ and $C_T$ are the covariance matrices of the source and target data respectively, and $d$ is the dimension of the deep layer activation.

$S$ and $T$ are set up similarly as for DeerDANN, except that $T$ also contains the training set.
$T$ is constructed in this manner to resemble $S$ in class priors, but with the features of real deer instead of synthetic deer.
During training, source samples are used to compute a classification loss $\mathcal{L}_C$.
Further, a CORAL loss is computed using the source and target batch with Eq. (~\ref{CORAL_loss}).
The network is trained end-to-end using the composite loss
\begin{equation}
   \mathcal{L}_{\text{sum}} = \sum_{i \in S} \mathcal{L}_C^i + \lambda \cdot \mathcal{L}_{\text{CORAL}},
\end{equation}

\noindent where $\lambda$ is a hyperparameter controlling the trade-off between classification loss and CORAL loss.
By minimizing the composite loss $\mathcal{L}_{\text{sum}}$, the network learns to extract features with similar statistics from both the source and target domains, but with discriminative distributions for different classes. 
See Fig.~\ref{fig:architectures} for the overall DeerCORAL architecture.

\section{Experiments}

In our experiments we perform cropped bounding box classification of animal species in the CCT-20 data set.
Next to the CCT-20 images, we use a pool of 5k day and 5k night simulated deer images from where we sample subsets of synthetic deer for experiments.
All CCT-20 images and synthetic images are cropped to the provided bounding boxes and are rescaled to 299 $\times$ 299 pixels.

As baseline we use an Inception v3 model~\cite{szegedy2015} pre-trained on ImageNet~\cite{deng2009} with an initial learning rate of 0.0045, \mbox{RMSprop} with 0.9 momentum, and horizontal flipping, color jitter, and blur as data augmentation, following~\cite{beery2020}.
The Inception model is fine-tuned on the CCT-20 training set, as well as on the training set augmented with 10k synthetic deer images.
We refer to these models as \textit{Inception v3 real} (real samples only) and \textit{Inception v3 syn} (real and synthetic samples).

\subsection{Effect of Number of Simulated Deer on DeerDANN Performance}

For DeerDANN we use an ImageNet pre-trained ResNet50 model~\cite{he2015} without the last layer as feature extractor $F$.
Fully connected layers are used as discriminator $D$ (1024 - 1024 - 2) and as label classifier $C$ (1024 - 12).
During training we use an initial learning rate of $10^{-5}$ and the Adam optimizer~\cite{kingma2017} with L2 regularization.
Random cropping, color jitter, and horizontal flipping are applied on all real and synthetic samples as data augmentation.
The number of simulated deer is varied from 100 to 10k samples and models are selected using the trans+ validation set.

Additionally, we train a similar model with the same settings, where samples from all classes are sent to the discriminator instead of only deer samples.
We call this model AllDANN and use this model to investigate whether isolated domain adaptation on the deer class is favourable to domain adaptation on all classes.
For AllDANN the target domain also contains the training set, as shown in Fig.~\ref{fig:architectures}.

\begin{figure}[!htb]
   \centering
   \subfigure[DeerDANN]{\label{fig:b}\includegraphics[width=0.48\textwidth]{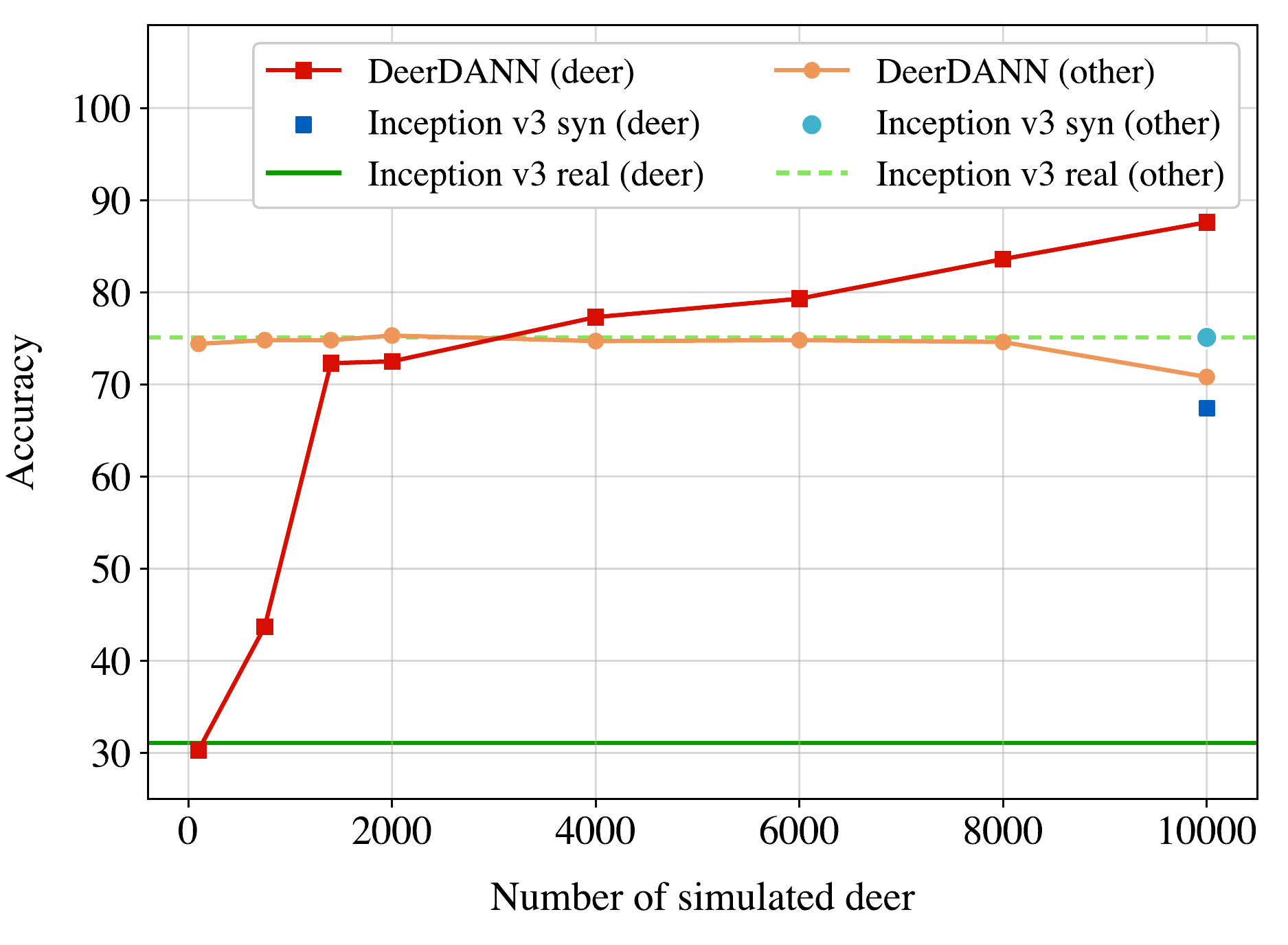}}
   \subfigure[AllDANN]{\label{fig:a}\includegraphics[width=0.48\textwidth]{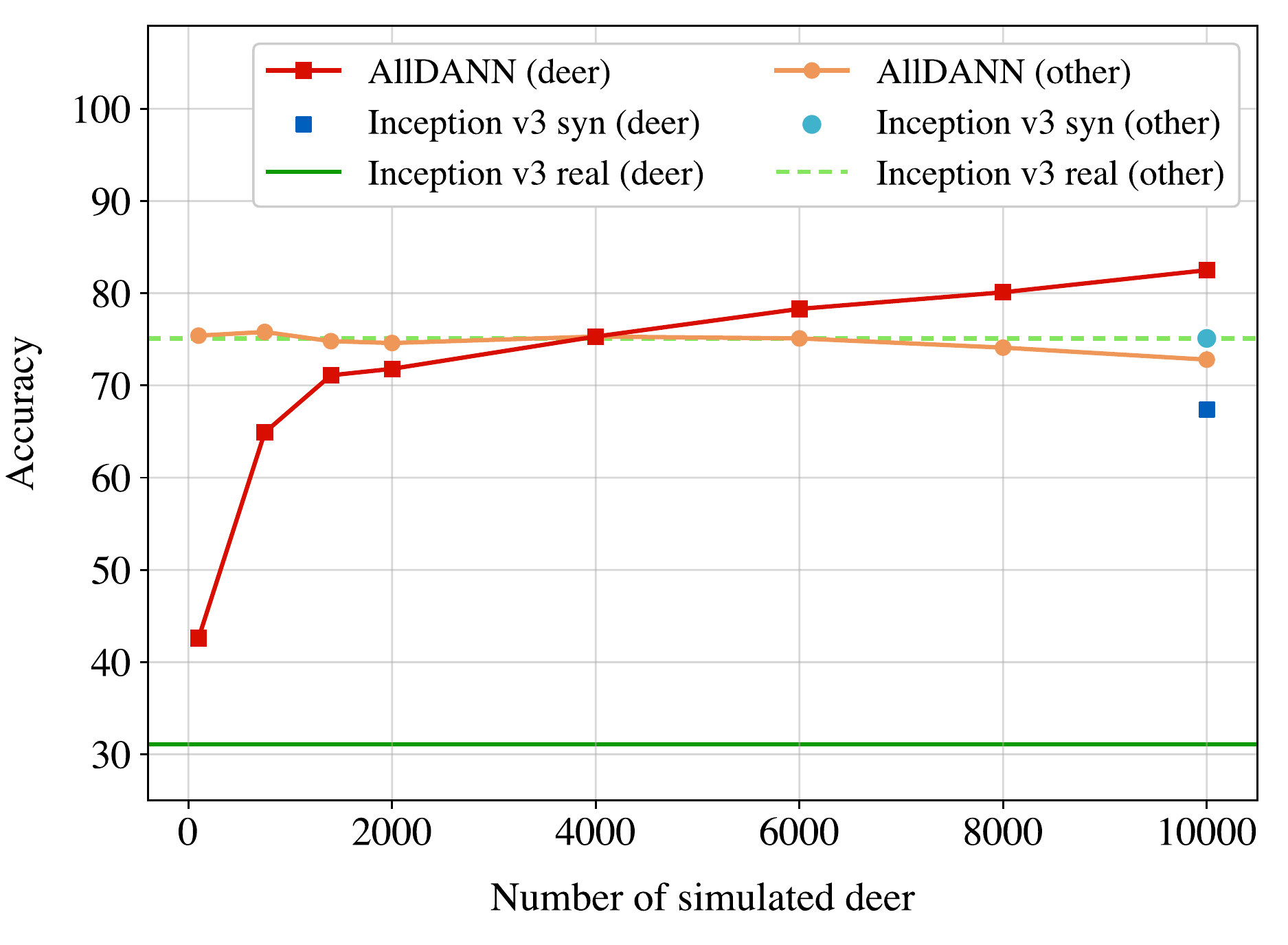}}
   \caption{Learning curves of DeerDANN and AllDANN for various numbers of simulated deer ranging from 100 to 10k. Accuracy is measured on the trans+ test set. For 1400 or more simulated deer, both methods have higher deer accuracies than the Inception v3 syn baseline.}
   \label{fig:comparison_DANN}
\end{figure}

For 1400 or more synthetic deer, both DeerDANN and AllDANN have higher accuracies than the Inception v3 syn baseline, trained with 10k synthetic deer, on the trans+ deer class (see Fig.~\ref{fig:comparison_DANN}).
The performance of both models for the other classes on average is similar to the baselines, for 8k or fewer simulated deer.
When adding up to 10k synthetic deer, the performance of the deer class increases considerably, at the cost of a larger drop in other classes.
In that case, the large deer accuracy increase is most likely due to the large deer class prior, which is four times larger than the prior of the second most common \textit{opossum} class.

Between DeerDANN and AllDANN, DeerDANN seems to perform better on the trans+ deer class for 4k or more simulated samples.
Thus, it seems beneficial to only channel the deer features to the discriminator instead of all features, when performing domain adaptation on a single class.

\subsection{Effect of Number of Simulated Deer on DeerCORAL Performance}

For DeerCORAL we again use an ImageNet pre-trained ResNet50 model as feature extractor and two fully connected layers as label classifier (1024 - 12).
Training is performed with an initial learning rate of $10^{-5}$, $\lambda$ equal to 0.5, Adam with L2 regularization and the same data augmentation techniques as for DeerDANN and AllDANN.
The last two layers are trained with a learning rate 10 times larger than the learning rate of the other layers, following~\cite{sun2016_2}.

\begin{figure}[!htb]
\centering
\includegraphics[width=0.5\linewidth]{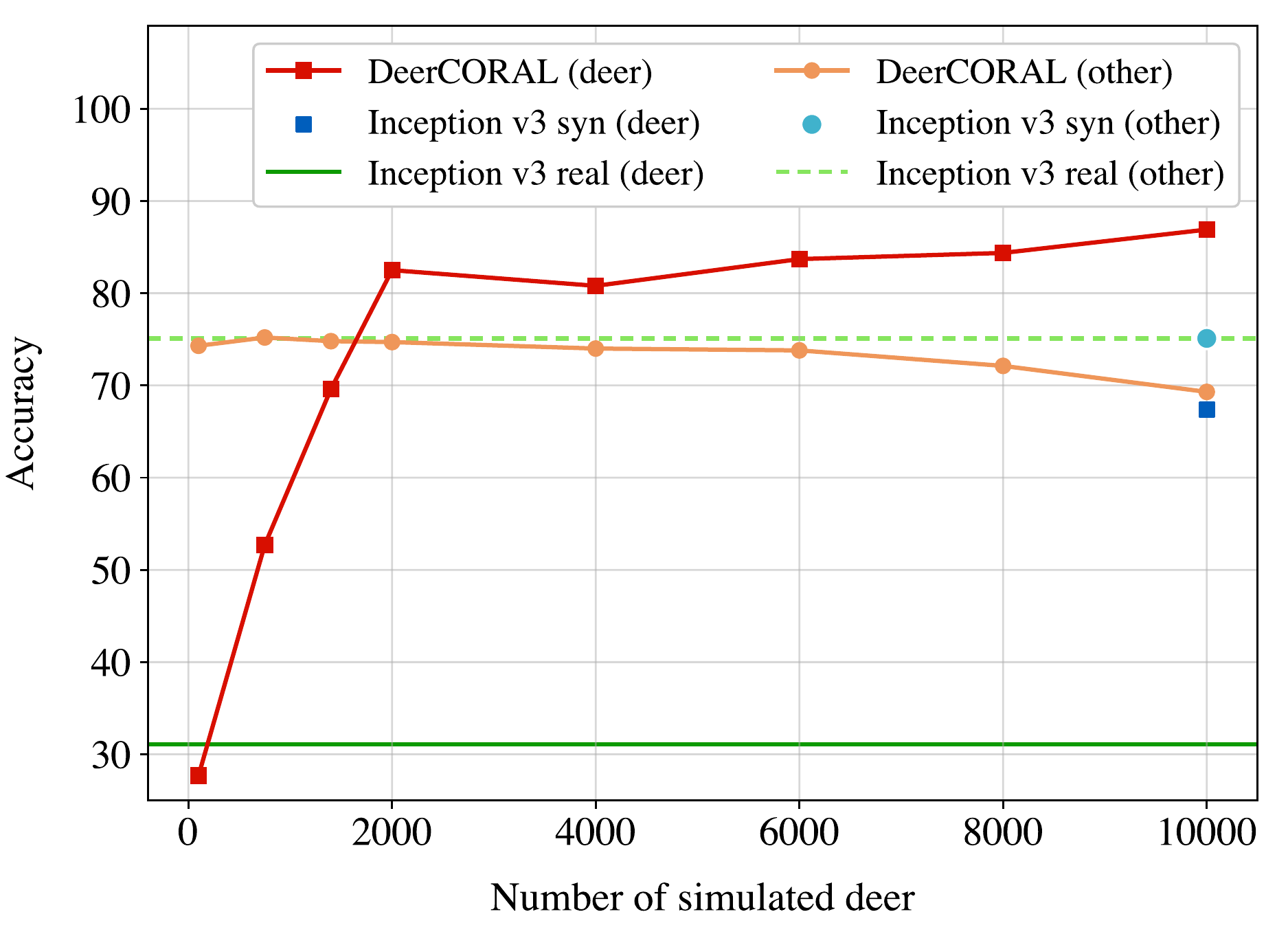}
\vspace{-1.6ex}
\caption{Learning curves of DeerCORAL for various numbers of simulated deer ranging from 100 to 10k. Accuracy is measured using the trans+ test set.
For 1400 or more simulated deer, DeerCORAL has higher deer accuracies than the Inception v3 syn baseline.}
\label{fig:comparison_CORAL}
\end{figure}

Similar to DeerDANN and AllDANN, DeerCORAL performs better than the baselines on the trans+ test deer class when using 1400 or more simulated deer (see Fig.~\ref{fig:comparison_CORAL}).
The average accuracy of other classes seems to drop faster than for the DANN models.
This drop can be avoided by using fewer deer samples, as DeerCORAL already shows significant improvement with 2k simulated deer.

\subsection{Comparison of DeerDANN, AllDANN, and DeerCORAL}

From each method we select the model with the highest deer accuracy and a loss of at most 1\% in trans+ other class accuracy.
These models are DeerDANN with 8k synthetic deer, AllDANN with 6k deer, and DeerCORAL with 2k deer.
We compare these models with the Inception v3 baselines, as well as an ImageNet pre-trained ResNet50 model, fine-tuned on the CCT-20 training set with 10k synthetic deer, which we call \textit{ResNet50 syn}.
ResNet50 syn is trained exactly like DeerDANN and AllDANN, and serves as an ablation model to study the performance of just the ResNet50 backbone.
We compare the model accuracies for the deer class as well as for the other classes on the cis and trans+ test sets (see Table ~\ref{tab:comparison}).

\begin{table*}[!htb]
\centering
\vspace{1ex}
\caption{A comparison between AllDANN (6k syn deer), DeerDANN (8k syn deer), DeerCORAL (2k syn deer), the Inception v3 baselines (0 and 10k syn deer), and an ablation ResNet50 model (10k syn deer) on the cis and trans+ test sets.}
\vspace{1ex}
\label{tab:comparison}
\resizebox{0.8\textwidth}{!}{%
\begin{tabular}{lcccc}
  \toprule
                       & trans+ deer & cis deer & trans+ other (avg.) & cis other (avg.) \\
                       \cmidrule{2-5} 
  Inception v3 real        & 31.1 & 51.7 & 75.1  & 89.5 \\
  Inception v3 syn         & 67.4 & 68.3 & 75.1 & 91.0 \\
  ResNet50 syn             & 42.3 & 58.7 & 74.1 & 90.0 \\
  AllDANN                  & 78.3 & 94.2 & 75.1 & 89.4 \\
  DeerDANN                 & 83.6 & 96.4 & 74.6 & 89.3 \\
  DeerCORAL                & 82.5 & 97.1 & 74.7 & 90.0 \\
  \bottomrule
\end{tabular}}
\end{table*}

First of all, Inception v3 syn performs better than ResNet50 syn in each category. 
Thus, the improvements of the domain adaptation models are not caused by the ResNet50 feature extractor.
Further, all domain adaptation models have higher deer accuracies than the Inception baselines.
DeerDANN has the largest accuracy improvement in the trans+ deer class of 24.0\% versus 16.2\% improvement of AllDANN and 22.4\% of DeerCORAL.
For the cis deer class DeerCORAL performs best, with 42.2\% improvement versus 37.9\% and 41.1\% of AllDANN and DeerDANN respectively.
All models perform quite similarly for the other trans+ classes, but slightly less good on the other cis classes.
All domain adaptation models require fewer than 10k deer to achieve higher deer accuracies than the Inception baselines, which were trained using 10k synthetic deer.
In particular, DeerCORAL requires few simulated deer samples (2k) versus 8k of DeerDANN and 6k of AllDANN.

\subsection{Network Feature Visualization}

\begin{figure}[!htb]
   \centering     %%% not \center
   \subfigure{\includegraphics[width=0.8\textwidth]{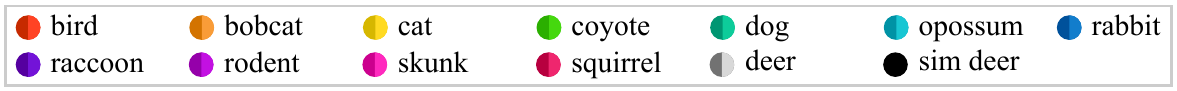}}\addtocounter{subfigure}{-1}
   \vspace{-1em}

%   \subfigure[AllDANN features]{\includegraphics[width=0.32\textwidth]{images/AllDANN_features.png}}
%   \subfigure[DeerDANN features]{\includegraphics[width=0.32\textwidth]{images/DeerDANN_features.png}}
%   \subfigure[DeerCORAL features]{\includegraphics[width=0.32\textwidth]{images/CORAL_features.png}}
   
  \subfigure[AllDANN features]{\includegraphics[width=0.32\textwidth]{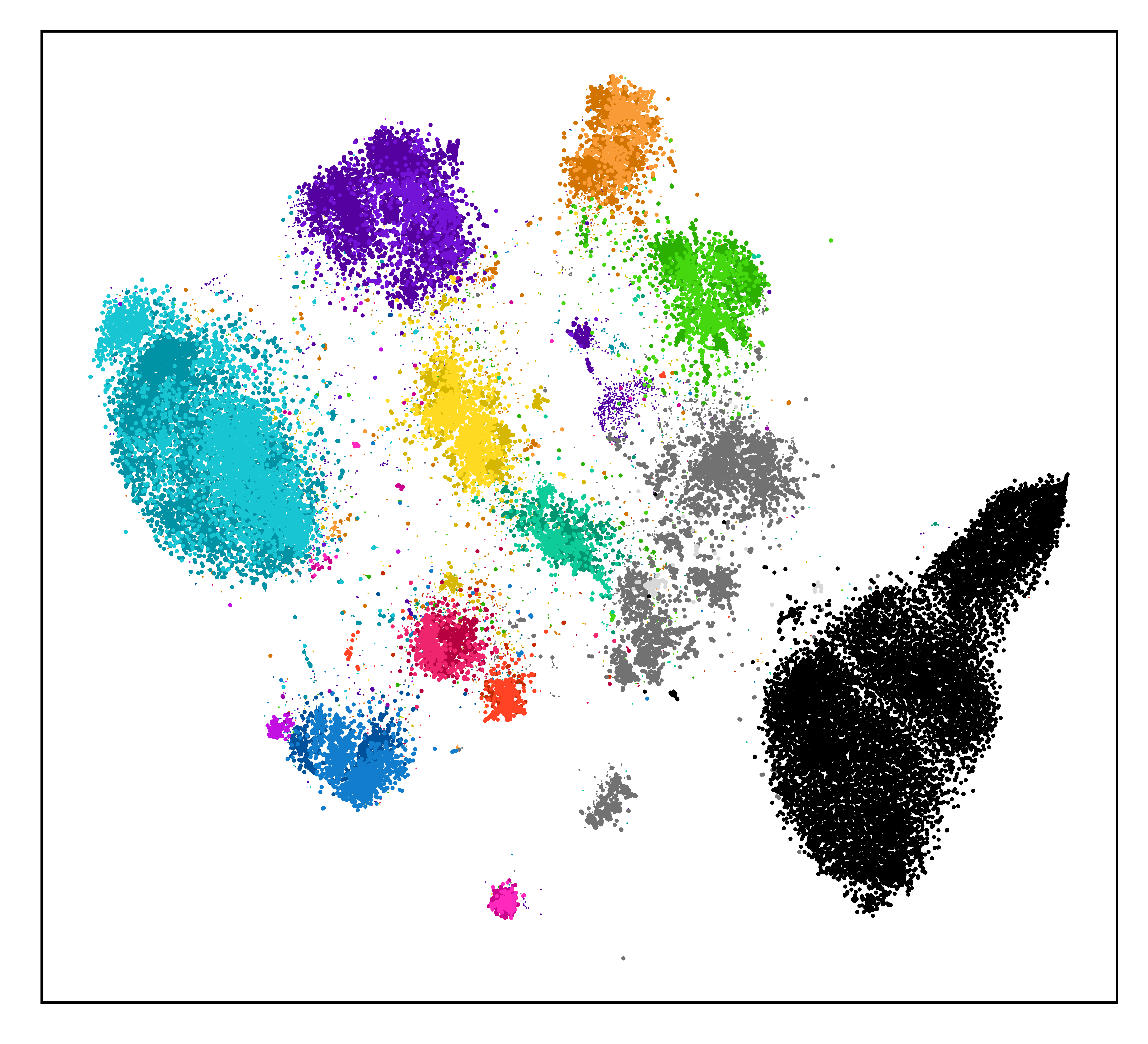}}
  \subfigure[DeerDANN features]{\includegraphics[width=0.32\textwidth]{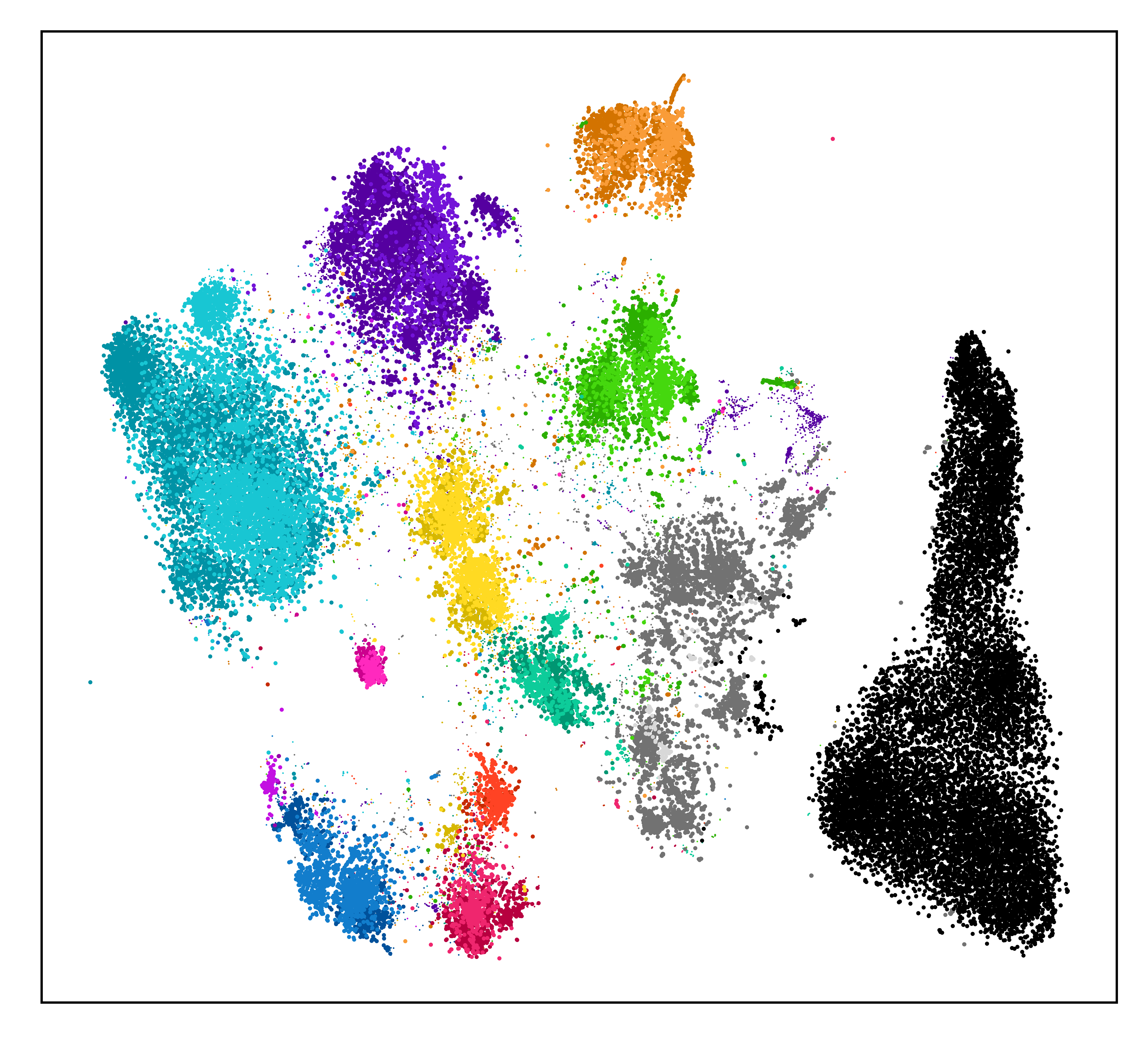}}
  \subfigure[DeerCORAL
  features]{\includegraphics[width=0.32\textwidth]{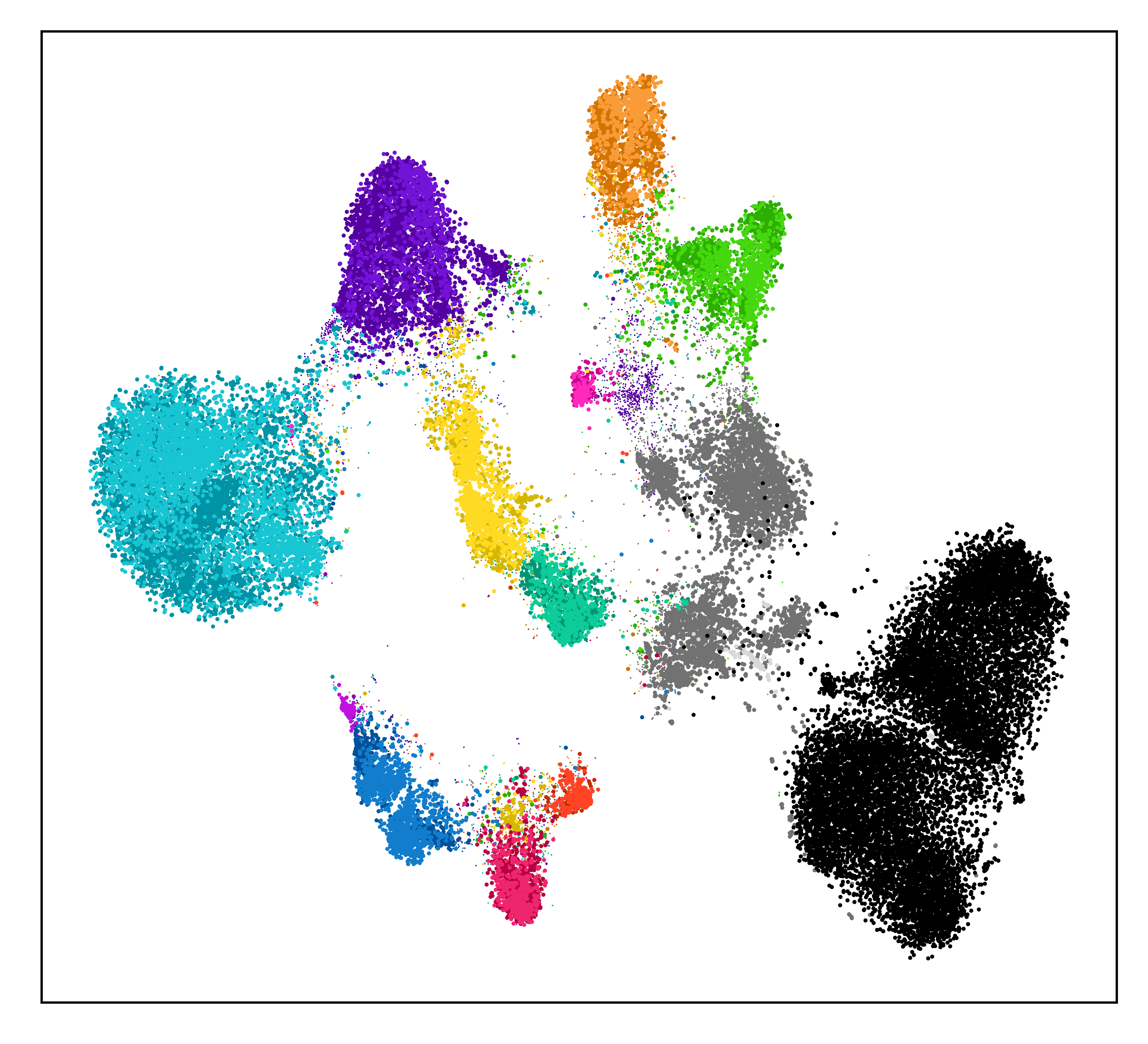}}

   \caption{Last pre-logit layer feature representations of all domain adaptation models of CCT-20 cis and trans+ test samples and 10k simulated deer. The simulated deer (black) and real deer (gray) are not clustered together, which indicates that the models learn the deer class bimodally. Small and large dots respectively show incorrect and correct classifications.
   Different hues of the same color represent trans+ samples (dark) and cis samples (light).}
   \label{fig:visualization}
\end{figure}

In~\cite{beery2020}, the CCT-20 classifier learned the deer class bimodally, as the network features of real and synthetic deer did not overlap.
Our expectation is that features generated by our models are similar for the real and synthetic deer samples.
We perform 200-dimensional PCA and t-SNE~\cite{tsne2008} afterwards on the feature activations of the last pre-logit layer to visualize the features, as seen in Fig.~\ref{fig:visualization}.

Surprisingly, DeerDANN and AllDANN appear to cluster the real and synthetic deer separately.
Even when a discriminator is applied, the classifiers still learn the real and synthetic deer bimodally.
DeerCORAL also learns the real and simulated deer bimodally, which is not surprising as DeerCORAL only minimizes the second-order statistics between the source and target features.
Even though the real and synthetic features do not overlap, applying domain adaptation still improves the classification of the rare deer class.

\section{Discussion}

% Summary
We present two different methods of applying domain adaptation to improve the classification accuracy of a rare class using simulated samples of that class.
The first method (DeerDANN), based on the Domain-Adversarial Neural Network (DANN), incorporates a discriminator in a classifier to generate domain-invariant features.
We modify the DANN method such that only samples of the rare class are sent to the discriminator, to perform domain adaptation on the rare class only.
The second method (DeerCORAL) incorporates a correlation alignment loss in a classifier, to align the second-order statistics of the source and target domain.

% Conclusions
From our experiments we conclude the following points.
First of all, both DeerDANN and DeerCORAL have higher deer classification accuracies and similar average accuracies for other classes, when compared to our baseline network.
Second, DeerDANN has the largest improvement in deer classification in unseen locations, which justifies sending only deer samples to the discriminator.
Third, both models require fewer simulated samples to achieve considerably better deer accuracies than the baseline. 
DeerCORAL especially only requires 2k samples versus 10k samples used by the baseline.

Regardless of the improved deer classification using domain adaptation, the network features of the real and synthetic deer surprisingly still do not overlap when visualizing the features in two dimensions.
The question remains whether features of synthetic and real samples of a single rare class can be made to overlap using different domain adaptation techniques or other methods.

\section*{Acknowledgements}

Computational resources were provided by Microsoft AI for Earth.

\vskip 0.2in
\bibliography{sample}

\end{document}